\documentclass[acmtog]{acmart}

\usepackage{booktabs} 

\usepackage{enumitem}
\usepackage{colortbl} 

\usepackage[ruled]{algorithm2e} 

\SetAlFnt{\small}
\SetAlCapFnt{\small}
\SetAlCapNameFnt{\small}
\SetAlCapHSkip{0pt}

\acmConference[SIGGRAPH Asia’24]{SIGGRAPH Asia 2024}{December 3 - 6  2024}{Tokyo, Japanese}

\acmDOI{3680528.3687688}

\begin{document}
\title{Anim-Director: A Large Multimodal Model Powered Agent for Controllable Animation Video Generation}

\author{Yunxin Li}
\affiliation{
\institution{Harbin Institute of Technology, Shenzhen}
\country{China}}
\email{liyunxin987@163.com}
\author{Haoyuan Shi}
\affiliation{
\institution{Harbin Institute of Technology, Shenzhen}
\country{China}}
\email{g1016015592@gmail.com}
\author{Baotian Hu}
\authornote{Corresponding authors: Baotian Hu and Longyue Wang. \\
Contact Information: Yunxin Li, e-mail: liyunxin987@163.com; Baotian Hu, e-mail: hubaotian@hit.edu.cn; Longyue Wang, e-mail: vincentwang0229@gmail.com\\
Project: ~\url{https://github.com/HITsz-TMG/Anim-Director}}
\affiliation{
\institution{Harbin Institute of Technology, Shenzhen}
\country{China}}
\email{hubaotian@hit.edu.cn}
\author{Longyue Wang}
\affiliation{Shenzhen \country{China}}
\author{Jiashun Zhu}
\author{Jinyi Xu}
\affiliation{
\institution{Jilin University}
\country{China}}
\author{Zhen Zhao}
\affiliation{
\institution{Shanghai AI Lab}
\country{China}}
\author{Min Zhang}
\affiliation{
\institution{Harbin Institute of Technology, Shenzhen}
\country{China}}

\renewcommand\shortauthors{Yunxin Li, Haoyuan Shi, Baotian Hu, et al}

\begin{abstract}

Traditional animation generation methods depend on training generative models with human-labelled data, entailing a sophisticated multi-stage pipeline that demands substantial human effort and incurs high training costs. Due to limited prompting plans, these methods typically produce brief, information-poor, and context-incoherent animations. To overcome these limitations and automate the animation process, we pioneer the introduction of large multimodal models (LMMs) as the core processor to build an autonomous animation-making agent, named \textbf{Anim-Director}. This agent mainly harnesses the advanced understanding and reasoning capabilities of LMMs and generative AI tools to create animated videos from concise narratives or simple instructions. Specifically, it operates in three main stages: Firstly, the Anim-Director generates a coherent storyline from user inputs, followed by a detailed director’s script that encompasses settings of character profiles and interior/exterior descriptions, and context-coherent scene descriptions that include appearing characters, interiors or exteriors, and scene events. Secondly, we employ LMMs with the image generation tool to produce visual images of settings and scenes. These images are designed to maintain visual consistency across different scenes using a visual-language prompting method that combines scene descriptions and images of the appearing character and setting. Thirdly, scene images serve as the foundation for producing animated videos, with LMMs generating prompts to guide this process. The whole process is notably autonomous without manual intervention, as the LMMs interact seamlessly with generative tools to generate prompts, evaluate visual quality, and select the best one to optimize the final output. To assess the effectiveness of our framework, we collect varied short narratives and incorporate various Image/video evaluation metrics including visual consistency and video quality. The experimental results and case studies demonstrate the Anim-Director’s versatility and significant potential to streamline animation creation.
\end{abstract}

%
%


%
%


\begin{teaserfigure}
    \centering
    \includegraphics[width=0.98\textwidth]{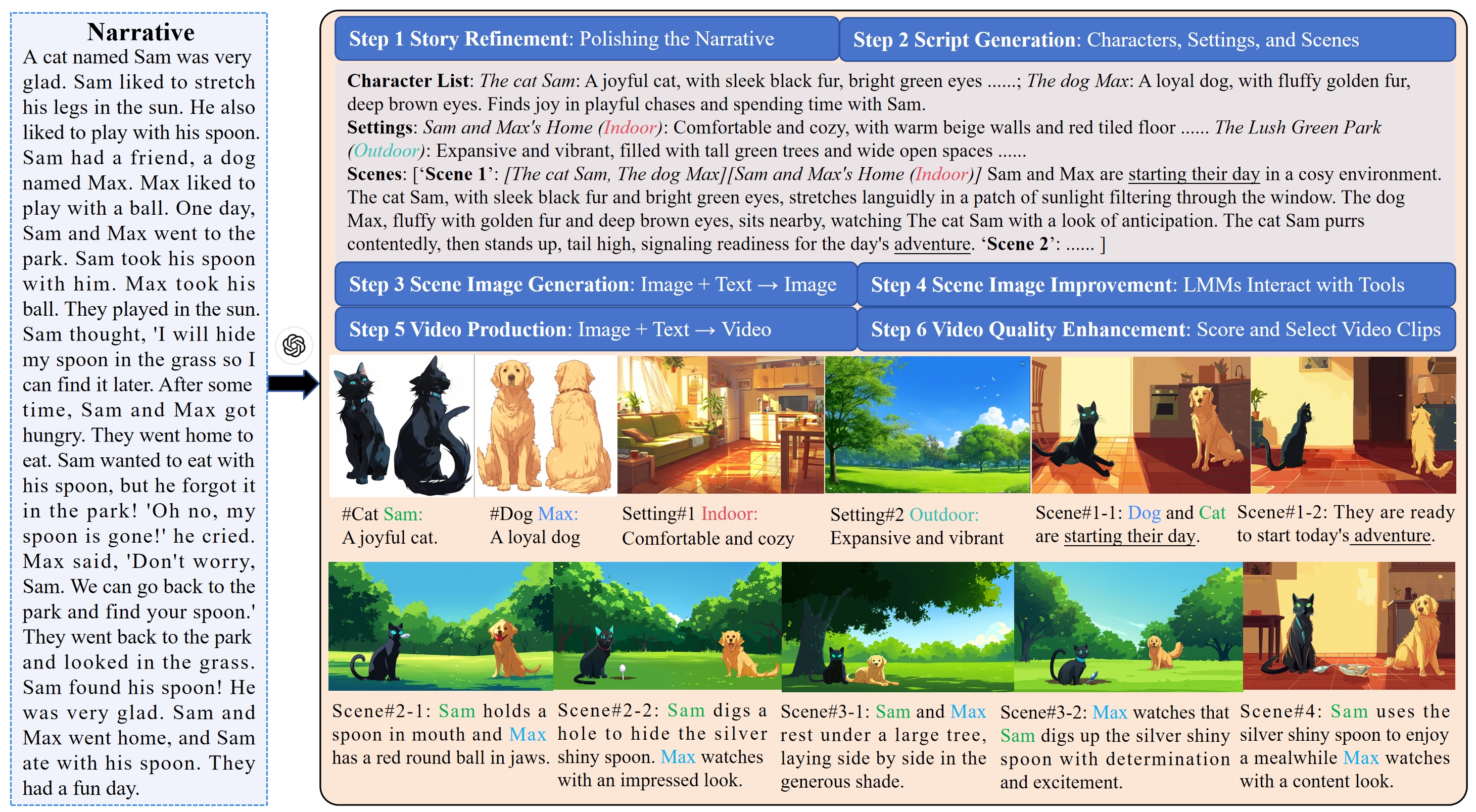}
    \caption{The visual examples of our Anim-Director. Given a narrative, it first polishes the narrative and generates the director's scripts using GPT-4. GPT-4 interacts with the image generation tools to produce the scene images through Image + Text → Image. Subsequently, the Anim-Director produces videos based on the generated scene images and textual prompts, i.e., Image + Text → Video. To improve the quality of images and videos, we realize deep interaction between LMMs and generative tools, enabling GPT-4 to refine, evaluate, and select the best candidate by self-reflection reasoning pathway.}
    \label{fig:intro-case}
\end{teaserfigure}

\maketitle

\section{Introduction}

The animation holds significant cultural, educational, and commercial value, impacting a wide range of industries from entertainment to marketing and education~\cite{tversky2002animation,wells2013understanding}.
The ongoing development of AI-driven generative tools~\cite{2danimation,wang20233d,azadi2023make} leads to new forms of entertainment and interaction, altering the paradigms of storytelling and its consumption. Moreover, these innovations possess the potential to democratize animation, making it more accessible to individuals and small teams without the resources of larger studios. Hence, a growing number of researchers are exploring AI techniques to automate the generation of animation videos.

Traditional AI-driven animation generation~\cite{Siyao_2021_CVPR,NEURIPS2022_48cca987,gong2023talecrafter} typically involves creating a series of images to depict a short story based on a text narrative and character portraits. The character portraits are often created by animation producers, a process that requires considerable manpower. Initially, the text narrative is converted into simple prompting sentences—commonly five to six in number. These sentences mainly convey relationships between characters and fundamental actions but often fail to maintain contextual semantic coherence throughout the story. Secondly, these prompting sentences, along with the portraits, are used to generate image sequences and animation videos~\cite{kumari2023multi,Rahman_2023_CVPR}, employing advanced image~\cite{NEURIPS2019_1d72310e,zhang2023texttoimage,Ge_2023_ICCV,yang_cvpr_2023,NEURIPS2023_821655c7} and video generation techniques, such as Image and Video diffusion models~\cite{NEURIPS2023_2468f84a,diffusion_Pami}. Hence, the produced animation videos are often brief and contextually incoherent, due to the limitations inherent in the visual representations and the concise nature of the story cues.

Previous approaches to animation video generation often involve a sophisticated multi-stage pipeline that necessitates significant human effort, particularly in training multiple models on human-labeled animated data~\cite{Li_2019_CVPR,maharana2021integrating}.
For instance, the recently introduced TaleCraft~\cite{gong2023talecrafter}
trains two models to complete the whole story visualisation process, including Text-to-Layout (T2L), and Controllable Layout-to-Image (C-L2I). Make-A-Story~\cite{Rahman_2023_CVPR} presents an autoregressive diffusion-based framework with a visual memory module, which relies on extensive story data for effective training. These models trained on limited datasets have poor generalization on new stories and training multiple models is troublesome.
Fortunately, the advent of large multimodal models (LMMs)~\cite{yao2022react,qin2023toolllm} such as GPT-4V~\cite{gpt4} and Gemini~\cite{team2023gemini},
has significantly enhanced the scope of capabilities in processing extensive multimodal information and performing chain-of-thought (COT) reasoning~\cite{wei2023chainofthought}. 
These models have been leveraged to develop autonomous agents~\cite{wang2023survey_agents} that can interact with external tools~\cite{li2024visiongraph} in human-like manners, thereby enhancing workflow efficiency. 

Inspired by these advancements, our work introduces a pioneering effort to employ an LMM-driven agent, named \textbf{Anim-Director}, designed to assist in producing contextually coherent animation videos. The core insight of the Anim-Director is allowing an LMM to act as an autonomous director, orchestrating the entire animation-making process. Specifically, as shown in Figure~\ref{fig:intro-case}, it manages the animation creation workflow through several critical stages:
\begin{itemize}[leftmargin=*]

\item \textbf{Story Refinement}: Given a tiny story or short narrative, we first use LMMs to enhance and develop it into a detailed, coherent, and plot-rich story, e.g., introducing character dialogue and story details while retaining the characters’ names. This approach is effective in supporting the creation of content-rich animations.

\item \textbf{Script Generation}: Anim-Director creates a detailed and well-structured script from the expanded story. As depicted in the top part of Figure~\ref{fig:intro-case}, this script contains thorough descriptions of all characters and settings (interiors and exteriors), along with carefully planned shot divisions (Scene descriptions). Each scene contains the corresponding characters, settings, and descriptions. 

\item \textbf{Scene Image Generation}: Firstly, we utilize the image generation tool\footnote{Midjourney: \url{https://www.midjourney.com/}} to produce the initial visual images including characters and settings. It constructs detailed visualizations of characters, interiors, and exteriors from textual descriptions. These images constitute the foundational visual elements for scene construction, which are integrated with specific textual prompts to result in unified visual-language prompts for the image generator. This process follows a Text + Image → Image way, wherein LMMs craft the textual prompts tailored to the scene descriptions.

\item \textbf{Scene Image Improvement}: To improve the quality of scene images, we enable LMMs to perform a content consistency evaluation between generated images and scene descriptions. This evaluation process includes selecting the most accurate depiction from a set of four candidate images. Additionally, to maintain visual consistency of characters across scenes, we employ the SAM tool~\cite{kirillov2023segment} to partition the generated images into distinct regions. This segmentation allows the LMM to assess and, if necessary, replace characters in the image to better match the predefined appearance of the characters. It ensures a coherent visual representation throughout the narrative.

\item \textbf{Video Production}: Anim-Director generates directive video prompts that target specific actions depicted within each scene, using the respective scene images and descriptions. It also predicts the optimal parameters of video generation tool\footnote{Pika: https://pika.art/} based on the action dynamics and visual content, aiming to capture the intended intensity and movement within the scenes. Subsequently, the scene image and video prompts are fed into the video generator with the appropriate parameter settings to produce video content, following the Image + Text → Video framework.

\item \textbf{Video Quality Enhancement}: Initially, we employ distortion detection and subject/background consistency evaluation approaches to assess the visual quality and contextual coherence of videos. This process involves splicing videos from consecutive scenes that share the same backgrounds to evaluate the overall narrative coherence. Subsequently, the LMM selects the optimal video from the top-ranking candidates based on the fidelity of content alignment between the generated video sequences and their respective narrative descriptions.
\end{itemize}
Overall, by automating these intricate tasks, Anim-Director significantly streamlines the animation production process through seamless integration with external tools, enhancing creative processes and minimizing the necessity for routine human oversight. This approach leverages the comprehensive capabilities of LMMs to improve both the efficiency and quality of animation production.

To verify the effectiveness of our approach, we collect varied narratives from TinyStories~\cite{eldan2023tinystories} dataset and employ a set of evaluation metrics introduced in the TaleCraft~\cite{gong2023talecrafter} and VBench~\cite{huang2023vbench} to assess the image and video quality. Experimental results demonstrate that Anim-Director can generate long animation videos with enhanced storylines and superior background consistency than baselines. Our detailed case analysis also reveals that agents built using LMMs and generative tools hold significant potential to improve long video creation. 

Our main contributions are in two key aspects:
\begin{itemize}[leftmargin=*]
    \item To the best of our knowledge, our work represents the first application of LMMs in developing autonomous agents for animation video generation. We introduce Anim-Director, a novel agent designed to manage the entire animation-making process autonomously. This agent can produce extended animation videos from brief narratives without the need for training, functioning akin to a film director.
    \item Anim-Director controls the generation of images and videos by leveraging foreshadowing images alongside corresponding textual prompts, i.e., Image + Text → Image/Video. Additionally, we augment the synergistic interaction between LLMs and generative tools, allowing LMMs to refine, evaluate, and select content produced by generative tools. It significantly enhances the quality of generated visuals via the designed reasoning pathway.
\end{itemize}

\begin{figure*}[t]
    \centering
    \includegraphics[width=0.99\textwidth]{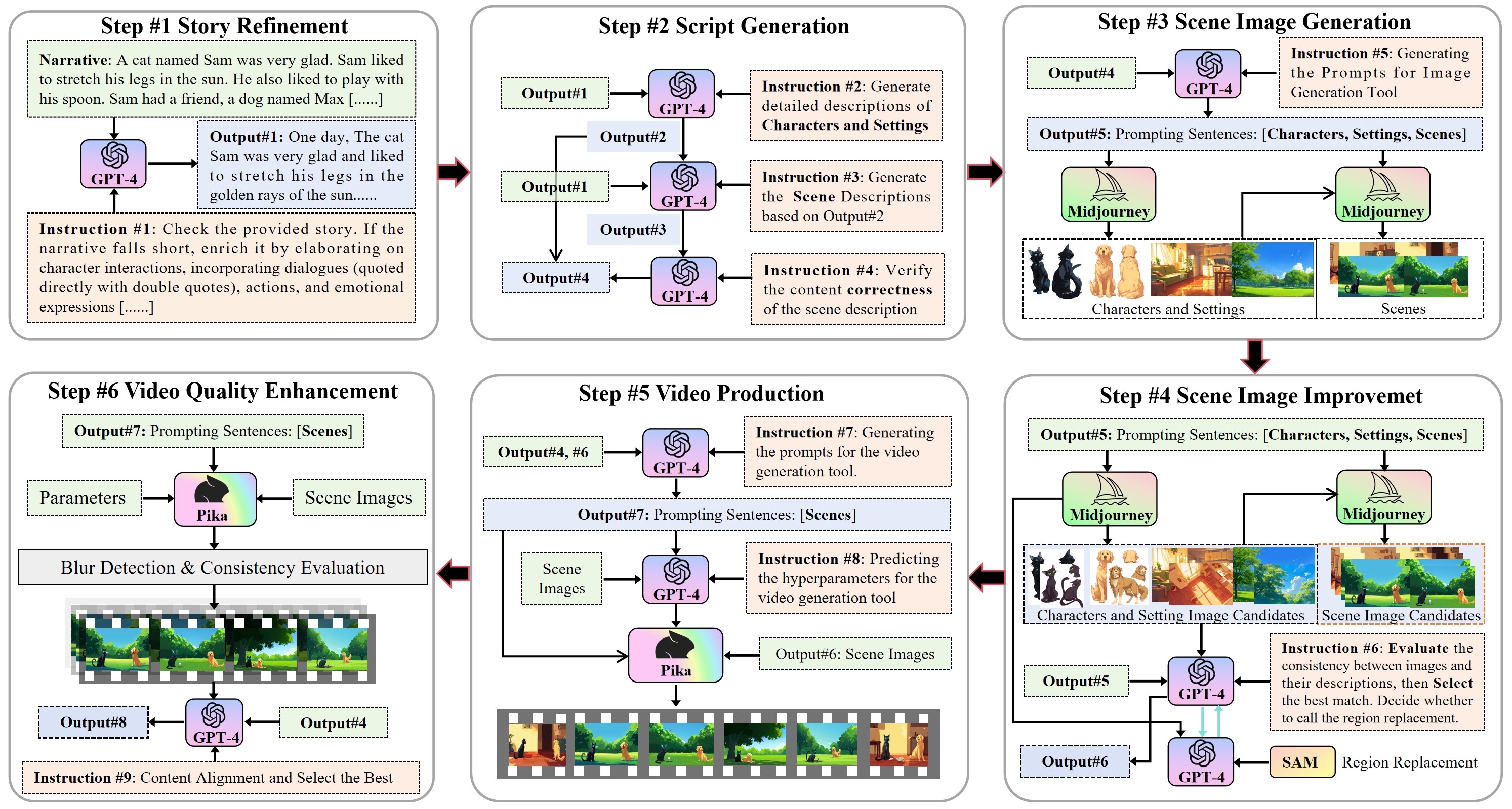}
    \caption{The overall workflow of our Anim-Director, which employs six steps to complete a whole animation video. The core technology of Anim-Director leverages the GPT-4 model as a director to execute a six-step automated management process, where we realize the deep interaction with generative tools. To enhance the quality of the generated content, we first use an `Image + Text → Scene Image/Video' approach for controllable visual content generation. We then apply designed enhancements to select the best images and videos from the candidates, ensuring superior output quality. }
    \label{fig:model-architecture}
\end{figure*}

\section{Related Work}

\subsection{Text-to-Animation Generation}

\textbf{Text-to-Video Generation}.
The field of text-to-video generation has witnessed significant advancements with the development of sophisticated models and techniques, greatly facilitating animation generation. These diffusion video generation models are initially trained on large-scale video-text datasets and can be fine-tuned with task-specific data to produce desired videos. Notable examples include ModelScope~\cite{wang2023modelscope}, Stable Video Diffusion~\cite{blattmann2023stablesvd}, VideoCrafter1~\cite{chen2023videocrafter1}, and VideoCrafter2~\cite{chen2024videocrafter2}. Additionally, models such as Dynamicrafter~\cite{xing2023dynamicrafter}, TaleCrafter~\cite{gong2023talecrafter}, and Vlogger~\cite{zhuang2024vlogger} exemplify the application of these techniques to generate high-quality animations. Tools like Gen-2~\cite{runawaygen2} and Pika further advance the field by offering powerful generalization capabilities, enabling the production of diverse video content.

\textbf{Text-to-Animation}.
Traditional animation generation methods mainly utilized story visualization technology, which involves segmenting stories and representing them through image generation technologies~\cite{chen2022character,Li_2019_CVPR,maharana2021integrating,maharana2021improving,Hei_Guo_Wang_Wang_Wang_Zhang_2024}. Initial approaches often employed Generative Adversarial Networks (GANs) \cite{goodfellow2014generative} or Variational Autoencoders (VAEs) \cite{doersch2016tutorial} for image creation. For example, StoryGAN \cite{Li_2019_CVPR} implements a dual-learning framework that utilizes video captioning to improve the semantic alignment between narratives and the resulting images. Furthermore, PororoGAN \cite{pororogan} enhances image relevance through the incorporation of sentence-level alignment and word-focused attention mechanisms, complemented by improved discriminators. 
Nevertheless, these approaches typically offer limited controllability in image generation and perform suboptimally on new or unfamiliar stories. 
Recent advancements in deep learning have led to the use of diffusion models for image and video generation, replacing GANs in story visualization. StoryDALL-E \cite{maharana2022storydall} uses latent diffusion models to generate images by aggregating data from current and preceding prompts via an autoregressive model. Similarly, Make-A-Story \cite{azadi2023make} employs an autoregressive, diffusion-based framework with a visual memory module, enhancing the continuity and relevance of visual narratives through a cross-attention mechanism within the feature space, building on AR-LDM \cite{pan2024synthesizing}.
The above methods require specific animation-related datasets to train their frameworks. These include the CLEVR-SV~\cite{Li_2019_CVPR}, Pororo-SV~\cite{pororogan}, CLEVR~\cite{johnson2017clevr}, Flintstones~\cite{maharana2021integrating}, and DiDeMo~\cite{anne2017localizing} datasets. However, models trained on these limited datasets often exhibit poor generalization to new narratives. Consequently, this paper proposes a framework driven by large models for animation generation, which does not rely on specific training datasets, allowing for adaptable application to new content without human intervention.


\subsection{Large Multimodal Models}
The rapid advancements in Natural Language Processing (NLP) have been driven by the development of Large Language Models (LLMs) such as Flan-T5 \cite{chung2022scaling}, GPT-4 \cite{gpt4}, and LLAMA \cite{touvron2023llama}. These models have expanded our language processing capabilities and paved the way for multimodal LLMs. Key factors include larger training datasets \cite{liu2023improved,li2024lmeye} and advancements in model architecture and design \cite{laurenccon2024obelics,yin2023survey}, which enhance the integration of neural networks and optimization techniques, improving language understanding tasks~\cite{chen-etal-2023-multi}.
Another emerging research area is the application of pre-trained LLMs to build multimodal LLMs, enhancing models for visually grounded tasks. Notable developments include LLaVA \cite{zhang2023llavar}, MiniGPT-4 \cite{zhu2023minigpt}, InstructBLIP \cite{dai2024instructblip}, Qwen-VL \cite{bai2023qwen}, Macaw-LLM~\cite{lyu2023macaw}, and LMEye \cite{li2024lmeye}, which can interpret images and respond to visual questions. These models typically use a pre-trained visual backbone \cite{clip} for image and video processing, an LLM \cite{touvron2023llama} for textual instructions and responses, and a cross-modal connector for effective visual-textual interaction. Compared to earlier multimodal models \cite{pororogan,diffusion_Pami}, the new multimodal LLMs inherit the reasoning and task adaptability of single-modality LLMs, making them suitable for diverse applications, including interactive generative tools for animation production.

\subsection{Large Models Powered Agents}
In the field of multimodal LLMs, the development of agents~\cite{wang2023survey_agents} powered by these models represents a significant advancement in artificial intelligence research. These agents effectively process multimodal inputs—text, images, and audio—driving significant research interest and solving complex tasks~\cite{wu2024perhaps}. Chameleon, a state-of-the-art framework introduced by \citet{lu2023chameleon}, exemplifies this progress. It integrates vision models, web search engines, Python functions, and rule-based modules to provide precise, updated responses, significantly enhancing natural language processing capabilities. \citet{toolformer} demonstrate that LMs can autonomously learn to utilize external tools through simple APIs, effectively combining robust computational power with versatile application.
Furthermore, \citet{wang2024mllmtool} unveiled MLLM-Tool, a system that marries open-source LLMs with multimodal encoders, enabling responsiveness to multimodal inputs. Additionally, \citet{wang2024mobile} introduces Mobile-Agent, an autonomous multi-modal mobile device that aids in retrieving useful applications and information. \citet{li2024visiongraph} develop a Description-Programming-Reasoning framework for the GPT-4V Agent, which performs multi-step reasoning on visual graphs through Python functions. These agents collectively showcase LLMs' potential in diverse applications, enhancing human creative processes. 

\section{Methodology}

\subsection{Overview}

We introduce Anim-Director, an autonomous agent powered by LMMs that leverages GPT-4 for generating contextually coherent animation videos. This agent integrates two generative tools: Midjourney for image generation and Pika for video production, enabling them to complete an entire animation project autonomously. As illustrated in Figure~\ref{fig:model-architecture}, Anim-Director operates through six main steps without being trained on specific datasets, allowing the LMM to comprehend and generate text, images, and videos, and select the best-generated content via a self-reflective reasoning process during interaction with the generative tools. Specifically, Steps \#1 and \#2 polish the input narrative and generate structured director-like scripts. Steps \#3 and \#4 focus on generating high-quality scene images by refining the output through multiple interactions between the LMMs and the image generation tool. Finally, Steps \#5 and \#6 involve producing the animation videos by invoking video generators, where the LMMs analyze the visual content of videos and select the best candidate from the generated videos.

\subsection{Story Refinement and Script Generation}

\textbf{Story Refinement}. Given a tiny story or short narrative, e.g., one topic sentence such as \textit{A cat and a dog are playing in the garden.}, we first need to refine it to be a plot-rich and detailed narrative that can support the content-rich animation generation such as introducing character dialogue and story details while retaining the characters' names. Hence, we design prompting Instruction\#1 as given in Table~\ref{script_generation} and enable GPT-4 to polish the initial inputs.

\textbf{Script Generation}. In this phase, our primary goal is to utilize GPT-4 to generate structured and detailed scripts, which will serve as blueprints for the subsequent creation of images and videos. The process unfolds in three main steps: 1) \textit{Character and scene setting extraction}: Following Instruction\#2, GPT-4 identifies characters and their settings within the story and unifies their formats. 2) \textit{Structured scene description generation}: Leveraging the extracted lists of characters and their respective settings, GPT-4 crafts a detailed description of each scene following Instruction\#3, focusing on the contextual coherence across scenes. 3) \textit{Verification and adjustment}: GPT-4 reviews the generated scene descriptions to ensure their relevance to the story. This involves checking the descriptions against the story content and the extracted character and scene settings, making necessary adjustments to maintain the integrity and coherence of the descriptions. The detailed Instructions are presented in Table~\ref{script_generation}.

\begin{table}[t]
    \centering
    \small
    \begin{tabular}{p{8.2cm}}
    \hline
     \cellcolor{cyan!15} Instruction\#1 (Step \#1 Story Refinement) \\
     \cellcolor{green!15} Ensure that the provided story \{Narrative\} is approximately 150 words and rich in plot details. If the story is lacking, enrich it by deepening character interactions, incorporating dialogues (enclosed in double quotes), detailing actions, and describing emotions. Enhance the plot’s continuity and coherence by making necessary adjustments, while retaining the original names of the characters. Moreover, consider introducing a pivotal moment or twist to increase engagement and narrative impact ......\\ 
    \hline
     \cellcolor{cyan!15} Instruction\#2 (Step \#2 Character and scene setting extraction) \\
     \cellcolor{green!15}Generate the character list and scene settings for the story \{Output\#1\} and present detailed descriptions of all characters and settings involved. \textbf{Character Descriptions}: Treat any man or animal appearing multiple times as a character. Each character should contain a comprehensive description that covers physical appearance, personality traits, and any significant actions or roles within the story...... \textbf{Setting Descriptions}: For each setting, use the following format: 'Name (Indoor/Outdoor):' followed by a brief introduction of the key elements, described in simple styles and colours. Specify whether each setting is indoor or outdoor ......\\ 
    \hline
    \cellcolor{cyan!15} Instruction\#3 (Step \#2 Structured scene description generation) \\
     \cellcolor{green!15} Generate the corresponding script with several scenes (including dialogues, actions, and emotions) using given detailed characters and settings \{Output\#2\}. Use the format ``[All Characters Included][Setting Included]: Description of this scene.'' to describe every scene. Select characters and settings from the provided lists and ensure that all characters and objects are named in their full form throughout the script ......\\ 
     \hline
     \cellcolor{cyan!15} Instruction\#4 (Step \#2 Verification and adjustment) \\
     \cellcolor{green!15} Check the generated scene descriptions. \textbf{1)}: Review each scene in Part 'Scenes', focusing on the characters and settings. If you find any character or setting not listed in Part 'Characters' or 'Settings', or any terminology inconsistency, respond with 'Yes.' \textbf{2)}: Ensure that every character who needs to appear in the scene's illustration is included in the initial character set. If any character is missing, respond with 'Yes.' If everything is correct in both steps, respond with 'No problem found.' If your answer is 'Yes.', provide a revised version of the script following demands: ......\\ 
     \hline
      \cellcolor{cyan!15} Instruction\#5 (Step \#3 Scene Image Generation) \\
     \cellcolor{green!15} 
     Generate prompts for the image generation tool Midjourney using the provided scripts \{Output\#4\}. These prompts should offer a clear visual description based on the content of the descriptions to aid in the creation of illustrations. Key instructions include:
     \textbf{1)}. All characters involved should be renamed following the rule below: [......] \textbf{2)}. For scene descriptions, simplify the actions and interactions between characters, avoiding overly complex or difficult-to-illustrate descriptions ......\\ 
    \hline
    \cellcolor{cyan!15} Instruction\#6 (Step \#4 Scene Image Improvement) \\
     \cellcolor{green!15} The image description is \{description\}. These images, numbered from 1 to 4, are produced based on this description. You need to answer: \textit{Which image most accurately reflects the setting described in the text?} Please provide a definitive answer using the format 'The answer is image x', followed by an analysis explaining your choice. For the selective images, you need further to check the character consistency between the scene and character images 5, following [......]. If not, generate a sign like ......\\ 
     \hline
    \end{tabular}
    \caption{Instructions of Story Refinement, Script Generation, Scene Image Generation, and Image Quality Improvement.}
   \label{script_generation}
\end{table}

\subsection{Scene Image Generation and Improvement}
In this stage, GPT-4 utilizes an image generator to generate images for each scene by the Prior Image + Text –> Scene Image approach and it enhances the quality through a self-reflection verification mechanism. The specific working process contains:

\textbf{Generate Scene Images}. First, we employ Midjourney to visualize characters and scene settings through the Text-to-Image method, as the dog and cat shown in Figure~\ref{fig:intro-case}. These images establish prior visual conditions for generating scene graphs, ensuring consistency in character and background depiction. Subsequently, we apply the Text + Image -> Image method to create specific visualization images for each scene, with GPT-4 generating prompts (Instruction \#5) tailored to each scene description according to the specific prompting styles of the image generator. Through these steps, we can obtain the initial images for each scene, which form the foundational visuals necessary for creating animation videos.

\textbf{Image Quality Improvement}. Considering the randomness of image generation, we introduce the self-reflection verification mechanism to gain high-quality images. It contains two steps: 1) Image generation tools first produce multiple candidates. We then utilize the visual-language understanding capability of GPT-4 to analyze them and select the best one that meets the specific demands, as shown in Instruction \#6, i.e., content is consistent with scene descriptions and image layout is reasonable. 2) We utilize GPT-4 to check the character consistency between the generated scene and prior characters' images. We employ the image segmentation tool SAM~\cite{kirillov2023segment} and the image region replacement capability of Midjourney to refine generated images. The outputs of this step are also fed into the previous step until passing the consistency inspection.

\subsection{Video Production and Quality Enhancement}
Similar to previous image generation, we also designed the video production and quality enhancement approaches directed by GPT-4. Instead of only using images or text as conditions to generate the video~\cite{chen2024videocrafter2}, we adopt a more controllable generation approach: Text + Image -> Video, where scene images and prompting text constrain the content of generated videos together.

\textbf{Video Production}. 
Initially, GPT-4 generates prompting sentences that describe the actions and emotions of characters, drawing from both the scene images and their descriptions. For the video generation model, the hyperparameter settings~\cite{azadi2023make,gong2023talecrafter} play an important role in generating high-quality videos. Hence, GPT-4 also predicts the most effective parameter settings based on the content of the prompting sentences and the visual cues from the scene images. These two elements—the prompting sentences and the parameter settings—are then input into the video generator to produce the final video product.
\begin{figure*}[t]
    \centering
    \includegraphics[width=0.99\textwidth]{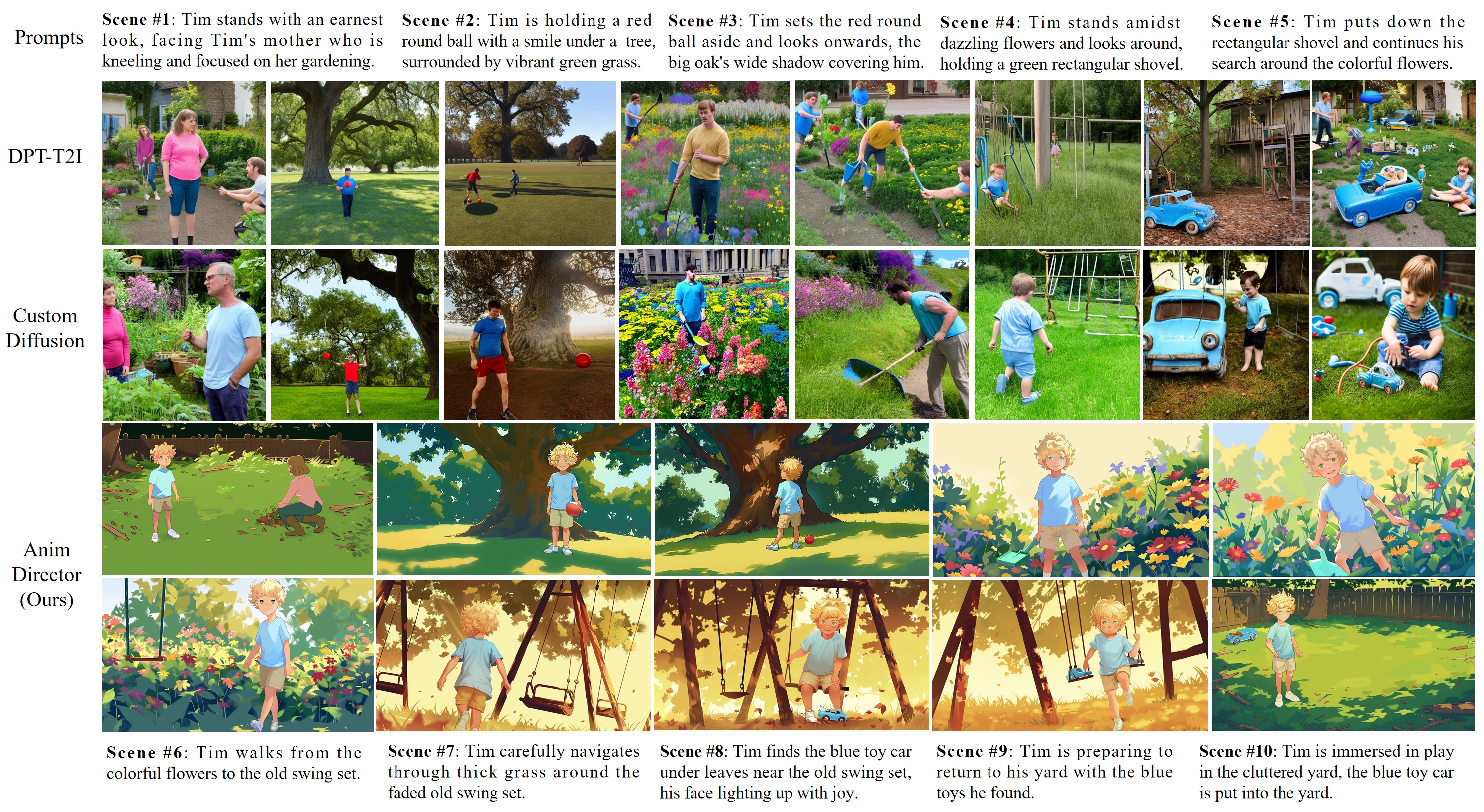}
    \caption{A comparative case showcasing various models. For DPT-T2I and Custom Diffusion, we display generated images for Scenes 1-5, 7-8, and 10. Images for all 10 scenes, including transition scenes 6 and 9, are presented for our model. Upon comparison, it is evident that our model exhibits superior visual coherence and quality. \textit{We convert the narrative into audio using Text-To-Speech (TTS), which is synchronized with the generated video. It is shown in supplementary materials.}}
    \label{fig:cases}
\end{figure*}

\textbf{Video Quality Enhancement}. 
Mostly video generation models like Stable Video Diffusion (SVD)~\cite{blattmann2023stable} and Pika often produce outputs that lack control in terms of visual detail, spatiotemporal consistency, and motion accuracy. To alleviate these issues, our approach involves generating ten initial video candidates. We then evaluate these candidates using video content metrics~\cite{huang2023vbench}, such as blur detection and consistency for both subject and background. From the top three candidates identified by these metrics, we use GPT-4 to select the best one based on how well the content aligns with scene descriptions, particularly in terms of action accuracy.
Finally, we splice all the scene videos to form the final complete animation video.
\begin{table}[t]
    \centering
    \small
    \begin{tabular}{p{8.2cm}}
    \hline
     \cellcolor{cyan!15} Instruction\#7 (Step \#5 Generating the prompts for video generation tool) \\
     \cellcolor{green!15} 
     Tou need to create simple visual animation descriptions based on the following: Image: shown in the input scene image; Animation description: \{Scene Description\}; Character description: \{Character\}. Your response should include:
     Part\#1. Screen Description; Part\#2. Action Description...... \\ 
    \hline
    \cellcolor{cyan!15} Instruction\#8 (Step \#5 Predicting the hyperparameter) \\
     \cellcolor{green!15} You need to predict suitable optimal parameter results by following the JSON format based on prompts and scene visualization, filling them with appropriate parameter information (option) Json Template
     ```
    \{
      "description": "Description of the scene",
      "option": \{
        "parameters": \{
          "motion": motion of scene actions or human expressions,
          "guidanceScale": Guidance scale for image generation,
          "negativePrompt": "Content to avoid generating"
        \},
        "camera": \{
          "zoom": "in"/"out"/null,
          "pan": "left"/"right"/null,
          "tilt": "up"/"down"/null,
          "rotate": "cw"/"ccw"/null
        \}
      \}
    \}
    ''' ......\\ 
     \hline
        \cellcolor{cyan!15} Instruction\#9 (Step \#6 Aligning Content and Selecting the Best) \\
     \cellcolor{green!15} You need to choose the better one from the three video candidates based on the following information. Each video is a composite image containing five temporal frames. The evaluation criterion is to see which video's displayed action is more in line with the description ......\\ 
     \hline
    \end{tabular}
    \caption{Video Production and Enhancement.}
   \label{video_generation}
\end{table}

\section{Experiments}

\subsection{Settings}
\textbf{Datasets}. For assessing the performance of the LMMs-driven animation generator, we compiled a dataset of 100 concise narratives sourced from TinyStories \cite{eldan2023tinystories}. 
These narratives are well-composed and grammatically near-perfect, spanning several paragraphs with a high degree of diversity. The stories feature interactions among animals, among humans, and between humans and animals, providing diverse dynamic scenarios for evaluation. 

\textbf{Comparing Models}. 
We primarily assess Anim-Director against other notable models in the field of animation and image generation. Specifically, we compare it with the open-source animation generation model Custom Diffusion~\cite{kumari2023multi} and other advanced image/video generation models. The DPT-T2I model introduced by~\citet{qu2024discriminative} enhances the discriminative capabilities of Text-to-Image (T2I) models, achieving more precise text-to-image alignment. Midjourney V6 is an AI-driven art generator that creates images from text descriptions and Pika v2 is a versatile video generator capable of dynamically modifying elements within the frame or changing styles on demand. VideoCrafter 2~\cite{chen2024videocrafter2} is a video generation model extended from Stable Diffusion, leveraging low-quality videos and synthesized high-quality images to obtain a high-quality video model. Moreover, we also compared our approach with some advanced video generation models to analyze the video quality, which contains Runaway-Gen2~\cite{runawaygen2}, ModelScope~\cite{wang2023modelscope}, VideoCrafter1~\cite{chen2023videocrafter1}, DynamiCrafter~\cite{xing2023dynamicrafter}, Stable Video Diffusion (SVD)~\cite{blattmann2023stablesvd} and AnimateDiff~\cite{guo2023animatediff}.

\textbf{Evaluation Metrics}.
Our experimental framework is designed to generate diverse, open-ended animations from short stories without references. Hence, we evaluate the model's capabilities in two primary areas: text-to-image generation and video quality.
\begin{itemize}[leftmargin=*]
    \item \textit{Text-to-Image}: We assess the alignment between the generated images and the input text, as well as the consistency among different images, similar to \citet{gong2023talecrafter}. Initially, we measure text-image similarity using the CLIP~\cite{clip} feature space to evaluate how well the images align with the text. Next, we use image-image similarity in the same feature space to examine the consistency of characters across images.
    \item \textit{Video Quality}: We use the comprehensive benchmark suite for Video Generative Models reported by~\citet{huang2023vbench} to assess the quality of generated videos. This includes evaluating temporal consistency (background and subject consistency) and frame-wise quality (visual distortion detection). Following \citet{gal2024breathing}, we also assess the text-video alignment score.
\end{itemize}

\begin{table}[t]
\centering
\setlength{\tabcolsep}{1.3pt} 
\renewcommand{\arraystretch}{1.10} 
\small
\begin{tabular}{lccc}
\hline
\textbf{Method}        & \textbf{Coherence} & \textbf{I-I Sim} & \textbf{T-I Sim} \\ \hline
\textbf{Custom-Diffusion}~\cite{kumari2023multi} & 0.74     & 0.66    & 0.28   \\
\textbf{DPT-T2I}~\cite{qu2024discriminative} & 0.75     & 0.65    & \textbf{0.29}   \\
\textbf{MidJourney-V6} [2024]  & 0.76   & 0.69     &   0.28  \\
\textbf{Anim-Director (Ours)}       & \textbf{0.87} $\uparrow11\%$       & \textbf{0.85} $\uparrow16\%$   & \textbf{0.29}    \\
\textbf{Ours w/o Image Improvement}  & 0.82    & 0.82     & 0.28    \\

\hline
\end{tabular}
\caption{Quantitative text-image evaluations on Contextual Coherence (Coherence), Image-Image Similarity (I-I Sim), and Text-Image Similarity (T-I Sim).}
\label{image_comparing}
\end{table}

\begin{table}[t]
\centering
\setlength{\tabcolsep}{1.5pt} 
\renewcommand{\arraystretch}{1.10} 
\small
\begin{tabular}{lccccc}
\hline
\textbf{Method}        & \textbf{V-Q} & \textbf{Subject} & \textbf{BackG} & \textbf{T-V} &  \textbf{Avg.\#Len}\\ \hline
\textbf{VideoCrafter1} \cite{chen2023videocrafter1} & 0.55 &  0.71 & 0.86 & 0.15 & 17.3\\
\textbf{VideoCrafter2}~\cite{chen2024videocrafter2} & 0.70 &  0.81 & 0.88 & 0.17 & 17.4\\
\textbf{TaleCrafter}~\cite{gong2023talecrafter} & 0.65 &  0.71 & 0.79 & 0.18 & 17.4\\
\textbf{ModelScope} \cite{wang2023modelscope} & 0.70 &  0.48 & 0.75 & 0.15 & 26.9\\
\textbf{AnimateDiff} \cite{guo2023animatediff} & 0.71 &  0.78 & 0.84 & 0.16 & 17.3\\
\textbf{SVD} \cite{blattmann2023stablesvd} & 0.50 &  0.81 & 0.90 & 0.16 & 15.1\\
\textbf{DynamiCrafter} \cite{xing2023dynamicrafter} & 0.72 &  0.82 & 0.88 & 0.18 & 21.7\\
\textbf{Vlogger} \cite{zhuang2024vlogger} & 0.72 &  0.80 & 0.87 & 0.17 & 21.6\\
\textbf{Gen-2} [2023] & 0.70 &  0.82 & 0.88 & 0.18 & 42.3\\
\textbf{Pika-v2} [2024] & 0.66     & 0.82    & 0.90 & 0.18 & 32.5   \\
\textbf{Anim-Director (Ours)}  &  \textbf{0.74}       & \textbf{0.86}   & \textbf{0.93} & 0.19 & 35.0   \\
\textbf{Ours w/o Video Enhancement}  & 0.67   & 0.84     & 0.91 & 0.18 & 35.0     \\
\hline
\end{tabular}
\caption{Quantitative video quality comparisons on Distortion detection (V-Q), Subject consistency (Subject), Background consistency (BackG), and  Text-Video alignment score (T-V). A higher score indicates better performance. Avg.\#Len refers to the average duration (second) of videos.}
\label{video_quality}
\end{table}

\begin{figure*}[t]
    \centering
    \includegraphics[width=1.0\textwidth]{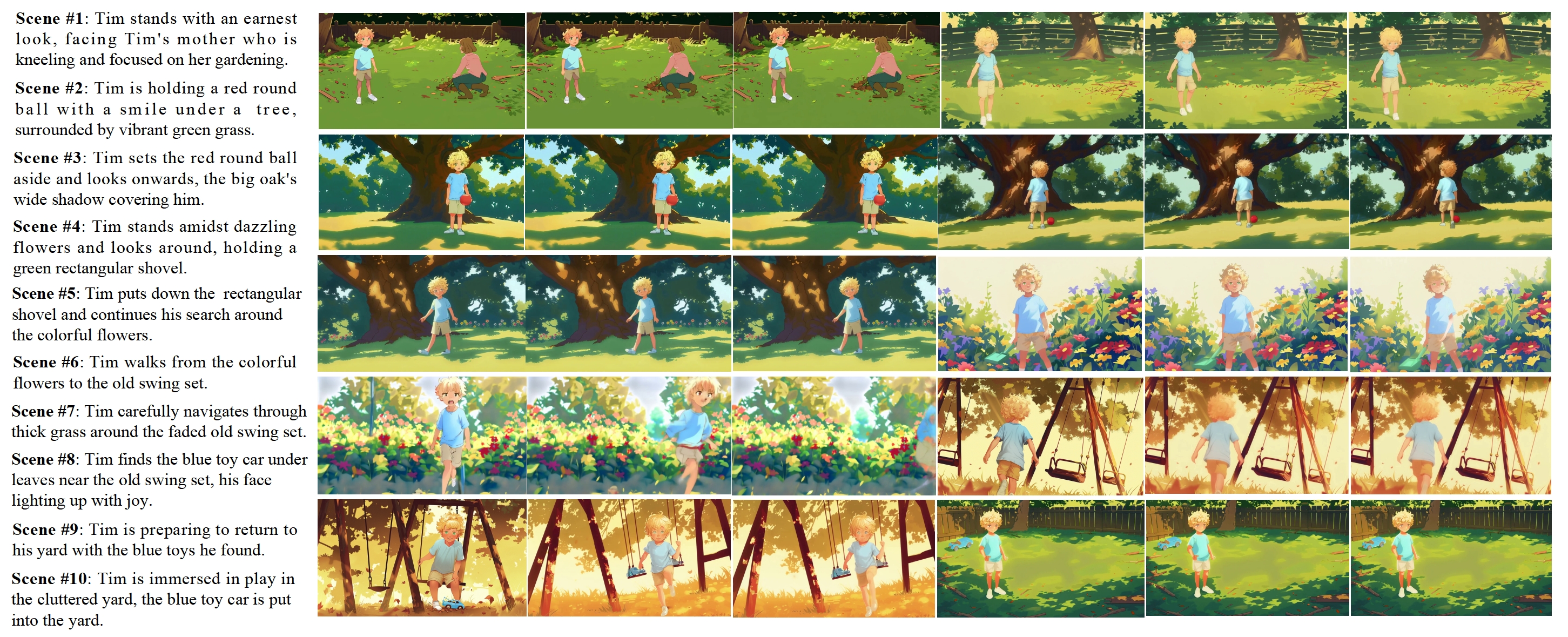}
    \caption{The extracted frames of videos generated by Pika.}
    \label{fig:cases-pika}
\end{figure*}

\subsection{Quantitative Comparisons}

From the experimental data presented in Tables \ref{image_comparing} and \ref{video_quality}, it is evident that our Anim-Director model achieves the highest performance in both image and video quality generation. Specifically, for contextual coherence detailed in Table \ref{image_comparing}, we adopt the average score of background and subject consistency from video quality evaluation. Our analysis indicates that the uniform performance of Text-Image similarity across various models can be attributed to their robust capabilities in generating images from textual descriptions, whereby they effectively capture essential information presented in the input. Furthermore, our model demonstrates superior performance compared to the most recent image generation models or generators, underscoring its effectiveness in story visualization. The results from ablation experiments (the last two rows) provide additional evidence that the image quality Improvement technique is effective. 

We have introduced video quality assessment into our experimental framework for the first time. The comparative results in Table \ref{video_quality} suggest that the proposed Anim-Director not only generates longer animated videos (3 seconds (s) per prompt compared to 1 second by Videocrafter2) but also maintains high visual quality with consistent subject and background elements. Moreover, the video enhancement technique we introduced elevates the overall quality of the whole video, which also indicates the capabilities of LMMs as reliable agents for long and plot-rich animation generation.

\subsection{Qualitative Comparisons}

\begin{figure*}
    \centering
    \includegraphics[width=0.98\textwidth]{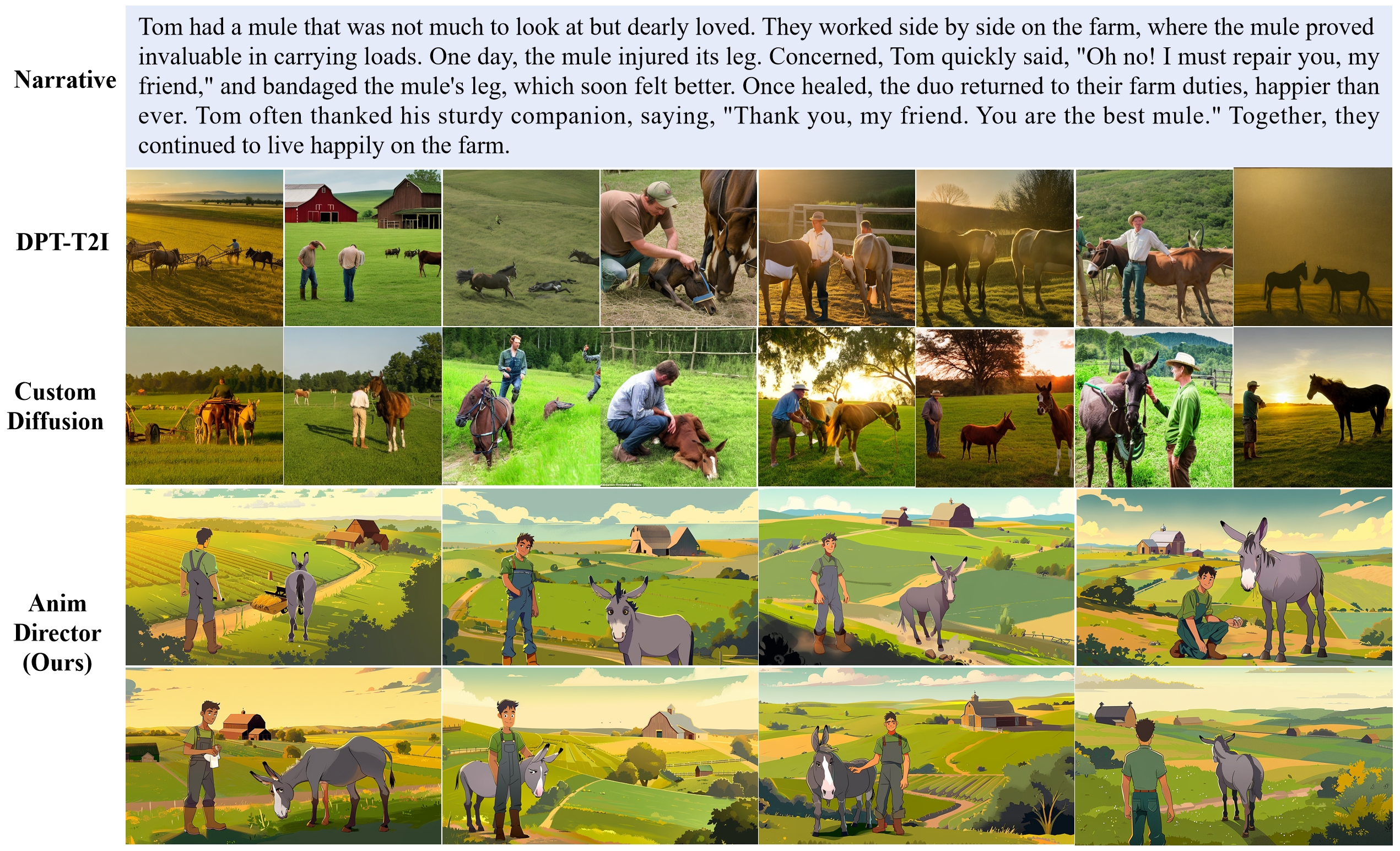}
    \caption{A case illustrating a story featuring human and animal interactions.}
    \label{fig:human-animal}
\end{figure*}

\begin{figure*}
    \centering
    \includegraphics[width=0.98\textwidth]{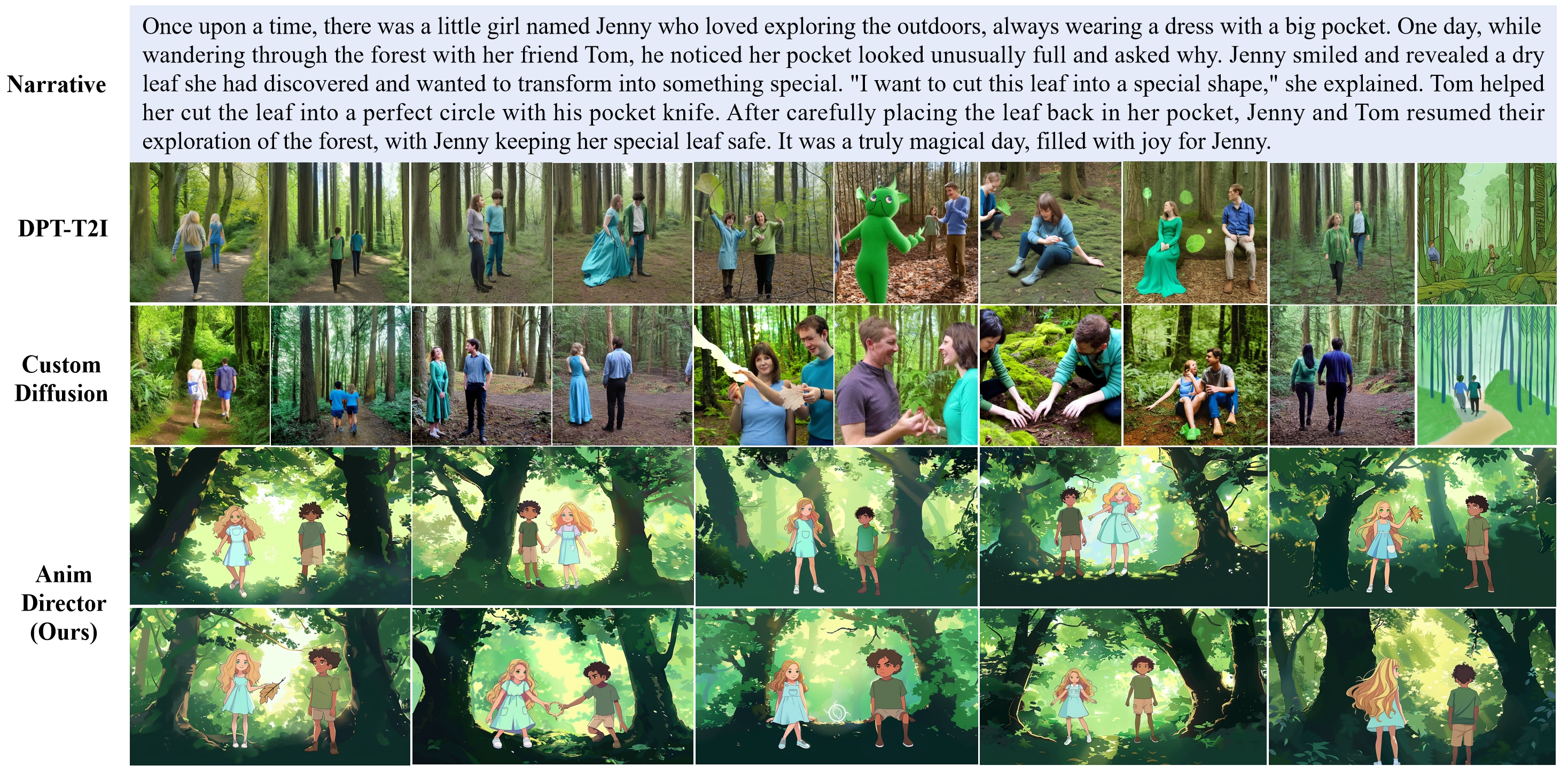}
    \caption{An illustrative case depicting a story involving human interaction.}
    \label{fig:human-human}
\end{figure*}

We present a comparative analysis in Figure~\ref{fig:cases} and the frames of videos generated by Pika (Figure \ref{fig:cases-pika}) to evaluate the generation quality of different models. Additional examples illustrating human-animal and human-human interactions are displayed in \textit{Figures \ref{fig:human-animal} and \ref{fig:human-human}}. Upon reviewing these cases, it becomes evident that our model consistently outperforms others in terms of visual coherence and quality, particularly in the rendering of human faces. Additionally, the consistency of humans, objects, and backgrounds in our results surpasses that of the baselines. Observing the generated images and videos shows that current video generation is difficult to control, especially for the consistency of different clips, and the background is easily changed.

\section{Limitations}
Our model primarily utilizes large multimodal model GPT-4 capabilities to interpret and analyze text, images, and videos. We employ generative tools such as Midjourney and Pika for producing images and videos. While these models represent the forefront of current technology, their generative processes are often not fully controllable and scene transitions may be choppy. This can result in the production of content that might be inappropriate, visually unappealing, or contextually inconsistent. In our forthcoming efforts, we aim to enhance the visual quality and contextual coherence of the generated videos, particularly across extended sequences.

\section{Conclusion}

This paper presents Anim-Director, an animation-making agent designed for generic animated video generation. Ours is the first work to leverage the powerful understanding, reasoning, and verification capabilities of Large Multimodal Models (LMMs) to produce long animated videos, allowing LMMs to interact with external generative tools deeply. The core insight of Anim-Director is enabling an LMM to function as an autonomous director, orchestrating the entire animation-making process. Previous methods rely on training generative models with human-labeled data, involving a complex multi-stage pipeline that requires substantial human effort and incurs high training costs. In contrast, our agent is flexible and training-free, capable of creating detailed animated videos from concise narratives or simple instructions. Additionally, we introduce a deep interaction approach between LMMs and external tools to enhance the quality of generated images and videos. We believe the proposed system is instructive and promising, with potential real-life applications in the future.



\section*{Acknowledge}
We would like to thank all reviewers for their detailed feedback.
This work is supported by grants: Natural Science Foundation of China (No. 62376067).

\bibliographystyle{ACM-Reference-Format}
\bibliography{sample-bibliography}



\end{document}